\documentclass[conference]{IEEEtran}
\IEEEoverridecommandlockouts
\usepackage{cite}
\usepackage{amsmath,amssymb,amsfonts}
\usepackage{graphicx}
\usepackage{textcomp}
\usepackage{xcolor}

\usepackage{algorithmic}
\usepackage{algorithm}
\usepackage{setspace}
\usepackage{booktabs}  
\usepackage{multirow} 
\usepackage{amsmath} 
\usepackage[T1]{fontenc}
\usepackage{bm}
\def\BibTeX{{\rm B\kern-.05em{\sc i\kern-.025em b}\kern-.08em
    T\kern-.1667em\lower.7ex\hbox{E}\kern-.125emX}}
\begin{document}

\title{Flow matching-based generative models for MIMO channel estimation\\}
\author{\IEEEauthorblockN{Wenkai Liu\IEEEauthorrefmark{1}, Nan Ma\IEEEauthorrefmark{1}, Jianqiao Chen\IEEEauthorrefmark{2}, Xiaoxuan Qi\IEEEauthorrefmark{1}, Yuhang Ma\IEEEauthorrefmark{1}}
\IEEEauthorblockA{\IEEEauthorrefmark{1}State Key Laboratory of Networking and Switching Technology\\
Beijing University of Posts and Telecommunications, Beijing 100876, China
\IEEEauthorblockA{\IEEEauthorrefmark{2}6G R\&D Department, ZGC Institute of Ubiquitous-X Innovation and Applications, Beijing 100876, China}
Email: \{liuwenkai, qixiaoxuan, mayuhang\}@bupt.edu.cn, jqchen1988@163.com}
Corresponding author: Nan Ma (manan@bupt.edu.cn)
}
\maketitle
\begin{abstract}
Diffusion model (DM)-based channel estimation, which generates channel samples via a posteriori sampling stepwise with denoising process, has shown potential in high-precision channel state information (CSI) acquisition. However, slow sampling speed is an essential challenge for recent developed DM-based schemes. 
To alleviate this problem, we propose a novel flow matching (FM)-based generative model for multiple-input multiple-output (MIMO) channel estimation.
We first formulate the channel estimation problem within FM framework, where the conditional probability path is constructed from the noisy channel distribution to the true channel distribution. In this case, the path evolves along the straight-line trajectory at a constant speed.
Then, guided by this, we derive the velocity field that depends solely on the noise statistics to guide generative models training. Furthermore, during the sampling phase, we utilize the trained velocity field as prior information for channel estimation, which allows for quick and reliable noise channel enhancement via ordinary differential equation (ODE) Euler solver. 
Finally, numerical results demonstrate that the proposed FM-based channel estimation scheme can significantly reduce the sampling overhead compared to other popular DM-based schemes, such as the score matching (SM)-based scheme. Meanwhile, it achieves superior channel estimation accuracy under different channel conditions.
\end{abstract}

\section{Introduction}
Equipped with a large number of antennas to utilize additional degrees of freedom in the spatial domain, multiple-input multiple-output (MIMO) technique can provide high data rates and spectral efficiency [1]. To obtain the above-mentioned benefits, accruate channel state information (CSI) is essential for supporting the subsequent signal processing, such as beamforming, channel equalization and signal detection [2]-[4]. Although the least squares (LS) channel estimator is widely used, it is becoming increasingly difficult to meet the requirements for precision due to its lack of robustness to nosie and interference [5].

Deep learning (DL) techniques offer a potential remedy by leveraging the inherent correlation of wireless channels. Studies [6]-[9] formulate the channel estimation task as an image restoration or denoising problem.
In [6] and [7], a channel recovery scheme was proposed where the channel matrix is first estimated via LS and treated as a low-resolution 2D image, and then it is enhanced through a super-resolution convolutional neural network. In [8], by exploiting spatial correlation and time correlation of different scenarios respectively, channel estimation is enhanced by deep convolutional neural network (CNN). Furthermore, the authors in [9] proposed deep residual learning based structure denoising CNN to perform the channel denoising task. These methods formulate the channel estimation problem as a supervised learning task aimed at reconstructing the ground-truth channel state. While having decent estimation performance, they require training the network on the large amount of labelled data, which is costly to acquire in real systems. Moreover, these methods are specific to measurement scenarios and usually lack the ability to generalize to wireless environments with varying signal-to-noise ratio (SNR).

The recent developed generative artificial intelligence (GAI) has revealed its potential to tackle persistent challenges in wireless communications [10].
Particularly noteworthy is the significant breakthrough achieved by learning complex data distributions and subsequently utilizing this prior information for channel denoising. 
In [11], the variational autocoder (VAE) is used as a generative prior for the estimation problem, and the derived estimator can be approximated as a minimum mean square error (MMSE) estimator.
In [12], the authors proposed using a Wasserstein Generative Adversarial Networks (GAN) to accurately learn the channel distribution. 
However, the effectiveness of these models was hampered by training instability, which prevented the generator from effectively learning from the critic's feedback and ultimately undermined the accurate reconstruction of the CSI [13].

Compared to VAE and GAN, the recent developed diffusion model (DM) can capture the intricate characteristics of data distributions, enabling the generation of content with rich, high-fidelity details. Furthermore, DMs are capable of producing more diverse images and have been demonstrated to be resistant to mode collapse [14]. Leveraging the advantages of generative diffusion models, the authors in [14]-[17] utilized diffusion models to capture the prior information of the channel, thereby proposing the channel denoising method analogous to diffusion probabilistic models. The study in [14] introduced a channel denoising diffusion model (CDDM) for semantic communication over wireless channels. In [15], a deep generative prior-assisted MIMO channel estimator based on the denoising diffusion generative model was proposed, which also handles quantized measurements when employing low-resolution ADCs. In [16], the posterior sampling method based on annealed Langevin dynamics was proposed. Although it leverages diffusion models to accurately estimate the score of the wireless channel, its performance advantage is offset by the computationally intensive sampling operations required at each diffusion step, leading to the significant delay in the channel estimation process. To accelerate estimation, a score matching (SM)-based variational inference method is introduced to directly approximate the posterior distribution of the MIMO channel, thereby achieving efficient channel recovery through reduced sampling steps [17].

However, despite existing research efforts to optimize the sampling efficiency of generative models, the number of steps required remains prohibitively high, presenting a great challenge for real-time channel estimation. Built upon continuous normalizing flows (CNFs), the flow matching (FM) framework represents a recent advancement in generative modeling, showcasing competitive or even superior results to DM methods in image synthesis and super-resolution [18]. Inspired by these advances, we propose a novel scheme for channel denoising that leverages FM framework. The main contributions of this paper are summarized as follows:
\begin{itemize}
    \item We formulate the channel estimation problem guided by FM framework. Specifically, the conditional probability path is constructed from the noisy channel distribution to the true channel distribution, which evolves along the straight-line trajectory at a constant speed. To the best of my knowledge, this is the first work of channel estimation within FM framework.
    \item Based on the above-mentioned path, we derive a velocity field that depends solely on the fixed noise statistics to guide the training of the generative model. By utilizing the converged velocity field as prior information for channel estimation, we can achieve fast and reliable enhancement over the noisy channel by Euler solver.
    \item Numerical simulations demonstrate that the proposed FM-based self-supervised channel estimation method achieves high estimation accuracy with fewer sampling steps compared to other popular DM schemes, and it has well generalization performance across different channel conditions.
\end{itemize}

\textit{Notation:} Symbols for vectors (lower case) and matrices (upper case) are in boldface. ${{\left( \cdot  \right)}^{H}}$ and ${{\left( \cdot  \right)}^{-1}}$ denote conjugate transpose (Hermitian) and inverse, respectively. $\mathbb{E}(\cdot )$ denotes expectation and ${{\left\| \Delta  \right\|}_{F}}$ denote frobenius norm for matrix $\Delta $.  $\mathcal{C}\mathcal{N}(\bm{\mu} ,\bm{\Sigma} )$ and $\mathcal{U}(a,b)$ denotes the Guassian distribution with mean $\bm{\mu} $ and covariance $\bm{\Sigma}$, and the uniform distribution between $a$ and $b$, respectively. $\frac{\partial }{\partial t}(\cdot )$ denotes the partial derivative with respect to time $t$.

\section{PRELIMINARIES AND PROBLEM FORMULATION}
\subsection{MIMO Channel Estimation}
We consider a MIMO communication system where the transmitter and receiver are equipped with \(N\) and \(M\) antennas, respectively. The wireless channel is characterized by the complex baseband channel matrix \(\mathbf{H} \in \mathbb{C}^{M \times N}\). For channel estimation, the transmitter sends the set of orthogonal pilot sequences over \(T\) time slots. For each pilot time slot \(t \in \{1, 2, \ldots, T\}\), the transmitted signal is \(\mathbf{p}_t \in \mathbb{C}^{N \times 1}\), whose elements are randomly selected quadrature phase shift keying (QPSK) symbols, and satisfy the orthogonality condition \(\mathbf{P} \mathbf{P}^H = E_p \mathbf{I}_N\), where \(\mathbf{P} = [\mathbf{p}_1, \mathbf{p}_2, \ldots, \mathbf{p}_T] \in \mathbb{C}^{N \times T}\) is the pilot matrix, and \(E_p\) is the total pilot power. The measurement vector $\mathbf{y}_t$ for the $t$-th time slot can be expressed as
\begin{equation}
\mathbf{y}_t = \mathbf{H} \mathbf{p}_t + \boldsymbol{\varepsilon}_t,
\end{equation}
where \(\boldsymbol{\varepsilon}_t \in \mathbb{C}^{M \times 1}\) is white Gaussian noise, following \(\mathcal{CN}(\boldsymbol{\varepsilon}_t \mid \mathbf{0}, \sigma^2 \mathbf{I}_M)\). Assuming the channel remains quasi-static over the consecutive \(T\) time slots, the system model can be compactly rewritten form as
\begin{equation}
\mathbf{Y} = \mathbf{H} \mathbf{P} + \mathbf{E},
\end{equation}
where \(\mathbf{Y} = [\mathbf{y}_1, \mathbf{y}_2, \ldots, \mathbf{y}_T] \in \mathbb{C}^{M \times T}\) is the measurement matrix, and \(\mathbf{E} = [\boldsymbol{\varepsilon}_1, \boldsymbol{\varepsilon}_2, \ldots, \boldsymbol{\varepsilon}_T] \in \mathbb{C}^{M \times T}\) is the noise matrix. Based on the above model, given the measurement matrix \(\mathbf{Y}\) and the known pilot matrix \(\mathbf{P}\), then we can estimate the channel \(\mathbf{H}\).
\subsection{Flow Model and ODE Solver}
We assume that the samples of the prior distribution ${{\mathbf{H}}_{0}}\sim {{p}_{0}}({{\mathbf{H}}_{0}})$ and the samples of the true channel distribution ${{\mathbf{H}}_{1}}\sim {{p}_{1}}({{\mathbf{H}}_{1}})$. The goal of the flow model is to transform the known prior distribution ${{p}_{0}}({{\mathbf{H}}_{0}})$ into the unknown true channel distribution ${{p}_{1}}({{\mathbf{H}}_{1}})$. To achieve this, we introduce a flow ${{\psi }_{t}}({{\mathbf{H}}_{t}}):={{\mathbf{H}}_{t}}\sim{{p}_{t}}({{\mathbf{H}}_{t}})$ dependent on the time variable $t\sim \mathcal{U}[0,1]$. Here, ${{p}_{t}}({{\mathbf{H}}_{t}})$ denotes the time-dependent probability density path with boundary conditions ${{p}_{0}}({{\mathbf{H}}_{0}})$ and ${{p}_{1}}({{\mathbf{H}}_{1}})$, whose evolution is described by the continuity equation:
\begin{equation}
\frac{\partial {{p}_{t}}({{\mathbf{H}}_{t}})}{\partial t}+\text{div}\left( {{p}_{t}}({{\mathbf{H}}_{t}}){{v}_{t}}({{\mathbf{H}}_{t}}) \right)=0,
\end{equation}
where $\text{div}\left({{p}_{t}}({{\mathbf{H}}_{t}}){{v}_{t}}({{\mathbf{H}}_{t}})\right)$ represents the divergence of the probability flow, and ${{v}_{t}}({{\mathbf{H}}_{t}})$ is the velocity field defined over the sample space and time. This velocity field governs the dynamic evolution of the probability density path ${{p}_{t}}({{\mathbf{H}}_{t}})$, which can be described by the continuous-time ordinary differential equation (ODE):
\begin{equation}
\frac{d{{\psi }_{t}}({{\mathbf{H}}_{t}})}{dt} = {{v}_{t}}({{\psi }_{t}}({{\mathbf{H}}_{t}})), \quad \text{where} \quad {{\psi }_{t}}({{\mathbf{H}}_{t}}) = {{\mathbf{H}}_{t}}.
\end{equation}
Given the vector field ${{v}_{t}}({{\mathbf{H}}_{t}})$, the ODE can be solved approximately via discretization. Here, we employ the first-order Euler method for sampling [16]. Specifically, let the current state be ${{\mathbf{H}}_{t}}$, the next state is computed iteratively as
\begin{equation}
{{\mathbf{H}}_{t+\Delta t}} = {{\mathbf{H}}_{t}} + \Delta t \cdot {{v}_{t}}({{\mathbf{H}}_{t}}).
\end{equation}
By setting the appropriate step size $\Delta t$, one can progressively approach the target distribution ${{p}_{1}}\left( {{\mathbf{H}}_{1}} \right)$ starting from the initial distribution ${{p}_{0}}({{\mathbf{H}}_{0}})$. In this case, the channel estimation problem can be formulated and solved within flow-based model framework.

\section{Proposed FLOW MATCHING-BASED CHANNEL ESTIMATION}
In this section, we first develop the conditional flow matching-based channel estimation framework to learn a velocity field as prior information, and then we provide the training process and sampling process of our proposed channel estimation method.
\subsection{Conditional Flow Matching for Channel Estimation}
The FM framework is grounded in learning a velocity field, as the true channel data distribution $p_1(\mathbf{H}_1)$ is unknown and the explicit form of the velocity field $v_t(\mathbf{H}_t)$ governing the probability path evolution is also inaccessible. The goal of FM is to train the parameterized neural network $u_t^\theta(\mathbf{H}_t)$ to approximate the true velocity field by minimizing the following loss function
\begin{equation}
   \mathcal{L}_{\text{FM}}(\theta) = \mathbb{E}_{t, p_t(\mathbf{H}_t)} \left[ u_t^\theta(\mathbf{H}_t) - v_t(\mathbf{H}_t) \
^2 \right], 
\end{equation}
where $\theta$ represents the learnable parameters of the velocity field $u_t^\theta(\mathbf{H}_t)$ . Flow matching constructs a probability path $p_t(\mathbf{H}_t)$ for $0 \le t \le 1$ connecting the known source distribution $p_0(\mathbf{H}_0)$ to the target data distribution $p_1(\mathbf{H}_1)$. The noisy CSI matrix is used as the starting point of the probability density path, i.e.,
\begin{equation}
    \mathbf{H}_0 = \mathbf{H}_1 + \tilde{\mathbf{E}},
\end{equation}
where $\tilde{\mathbf{E}}$ denotes the noise error matrix, distributed as $\tilde{\mathbf{E}} \sim \mathcal{CN}(0, \tilde{\sigma}^2 \mathbf{I}_M)$, $\tilde{\sigma }$ is the normalized standard deviation. This initialization strategy, where samples start near the target distribution, accelerates both training and sampling while simultaneously providing guidance for unconditional sampling.

In practical applications, the explicit expression of the marginal velocity field \( v_{t}(\mathbf{H}_{t}) \) is often unknown, making it infeasible to directly train an optimal probability density path. To address this challenge, conditional flow matching (CFM) provides an effective alternative. The key idea is to shift the focus from directly modeling the velocity field of the marginal probability path \( p_{t}(\mathbf{H}_{t}) \) to learning the conditional velocity field \( \mathbf{u}_{t}(\mathbf{H}_{t} | \mathbf{H}_{1}) \), which corresponds to a more tractable conditional probability path \( p_{t}(\mathbf{H}_{t} | \mathbf{H}_{1}) \).
By optimizing the CFM loss function \( \mathcal{L}_{\text{CFM}}(\theta) \), one can equivalently obtain the same optimal solution as marginal FM, i.e., ${{\mathcal{L}}_{\text{FM}}}(\theta )={{\mathcal{L}}_{\text{CFM}}}(\theta )$, the gradients of the two objectives with respect to the model parameters \( \theta \) are identical. The losses are defined as follows:
\begin{align}
    \mathcal{L}_{\text{CFM}}(\theta) &= \mathbb{E}_{t, p_{t}(\mathbf{H}_{t} | \mathbf{H}_{1})} \left[ \| u_{t}^{\theta}(\mathbf{H}_{t}) - v_{t}(\mathbf{H}_{t} | \mathbf{H}_{1}) \|^{2} \right], 
\end{align}
where $t\sim \mathcal{U}[0,1]$, the sample \( \mathbf{H}_{t} \sim p_{t}(\mathbf{H}_{t} | \mathbf{H}_{1}) \). The true conditional velocity field \( v_{t}(\mathbf{H}_{t} | \mathbf{H}_{1}) \) is determined by the time derivative of the conditional probability path:
\begin{equation}
    v_{t}(\mathbf{H}_{t} | \mathbf{H}_{1}) = \frac{d}{dt} \psi_{t}(\mathbf{H}_{t} | \mathbf{H}_{1}),
\end{equation}
Here, \( \psi_{t}(\cdot | \mathbf{H}_{1}) \) defines the conditional flow transformation from the prior distribution to the specific target \( \mathbf{H}_{1} \). The conditional flow matching loss \(\mathcal{L}_{\text{CFM}}(\theta)\) serves as the straightforward regression objective for training a velocity field neural network that describes the instantaneous velocity of samples. Once the loss function converges, this trained velocity field is used to transform the prior distribution \(p_0(\mathbf{H}_0)\) into the true channel distribution \(p_1(\mathbf{H}_1)\) along the probability density path \(p_t(\mathbf{H}_t)\).
To concretize the above framework, a widely adopted choice is the Gaussian conditional probability path constructed based on optimal transport theory, with the form as follows:
\begin{equation}
p_t(\mathbf{H}_t | \mathbf{H}_1) = \mathcal{N}(\mathbf{H}_t | \mu_t(\mathbf{H}_1), \sigma_t(\mathbf{H}_1)^2 I), 
\end{equation}
where the mean function $\mu_t(\mathbf{H}_1)$ and the standard deviation function $\sigma_t(\mathbf{H}_1)$ are designed as linear interpolation functions of time:
\begin{equation}
\mu_t(\mathbf{H}_1) = t \mathbf{H}_1, \quad \sigma_t(\mathbf{H}_1) = 1 - (1 - \sigma_{\min}) t, 
\end{equation}
where $\sigma_{\min}$ is set sufficiently small. Accordingly, the corresponding conditional flow $\psi_t(\mathbf{H}_t | \mathbf{H}_1)$ can be constructed, which maps the prior point $\mathbf{H}_0$ to the point $\mathbf{H}_t$ on the conditional path at time $t$.
\begin{equation}
\begin{aligned}
\mathbf{H}_t &= \psi_t(\mathbf{H}_t | \mathbf{H}_1) = \mu_t(\mathbf{H}_1) + \sigma_t(\mathbf{H}_1) \cdot \mathbf{H}_0 \\
&= t \mathbf{H}_1 + (1 - (1 - \sigma_{\min}) t) \mathbf{H}_0.
\end{aligned} 
\end{equation}
where ${{\psi }_{t}}({{\mathbf{H}}_{t}}|{{\mathbf{H}}_{1}})$ is an affine transformation which maps to a normal ditribution random variable with mean ${{\mu }_{t}}({{\mathbf{H}}_{1}})$ and standard deviation ${{\sigma }_{t}}({{\mathbf{H}}_{1}})$. Based on the above flow transformation, the conditional velocity field with initial state \(\mathbf{H}_0\) and final state \(\mathbf{H}_1\) can be directly obtained by taking the derivative with respect to \(t\)
\begin{equation}
\begin{aligned}
v_t(\mathbf{H}_t | \mathbf{H}_1) &= \frac{d}{dt} \left( \psi_t(\mathbf{H}_t | \mathbf{H}_1) \right) \\
&= \frac{d}{dt} \left[ t \mathbf{H}_1 + (1 - (1 - \sigma_{\min}) t) \mathbf{H}_0 \right] \\
&= \mathbf{H}_1 - (1 - \sigma_{\min}) \mathbf{H}_0.
\end{aligned}
\end{equation}
For simplicity, we set \(\sigma_{\min} = 0\). Then, this true velocity field \(v_t(\mathbf{H}_t | \mathbf{H}_1)\) is independent of time \(t\) and the current state \(\mathbf{H}_t\), and is constant equal to the difference between the final state and the initial state, i.e., \(v_t(\mathbf{H}_t | \mathbf{H}_1) = \mathbf{H}_1 - \mathbf{H}_0\). A crucial insight is that \(-\tilde{\mathbf{E}} = \mathbf{H}_1 - \mathbf{H}_0\), which implies that the parameter optimization process of \(u_t^\theta(\mathbf{H}_t)\) directly depends on the estimation error \(\tilde{\mathbf{E}}\). Therefore, the statistical properties of \(\tilde{\mathbf{E}}\) directly affect the training dynamics of the model. In other words, the goal of the FM-based model \(u_t^\theta(\mathbf{H}_t)\) is to predict the noise error \(-\tilde{\mathbf{E}}\) as accurately as possible, i.e.,
\begin{equation}
\mathcal{L}_{\text{FM}}(\theta) = \mathbb{E}_{t, p_t(\mathbf{H}_t | \mathbf{H}_1)} \left[ \| u_t^\theta(\mathbf{H}_t) + \tilde{\mathbf{E}} \|^2 \right].
\end{equation}
Once the accurate velocity field model is obtained through training, high-quality samples conforming to the true channel distribution \(p_1(\mathbf{H}_1)\) can be efficiently generated by numerically solving the corresponding ordinary differential equation.
\subsection{Training and Sampling for MIMO Channel Estimation}

The core of the FM-based channel estimation
method lies in the evolution of a probability density path
parameterized by a neural network, which is trained with the loss function as shown in (14).
The implementation of this training procedure relies on a modified UNet architecture that deeply integrates the convolutional residual block (ResBlock) with the convolutional attention block (AttenBlock) in the encoder-decoder symmetric structure of classical UNet, as shown in Fig. 1.
\begin{figure}
    \centering
    \includegraphics[width=0.99\linewidth]{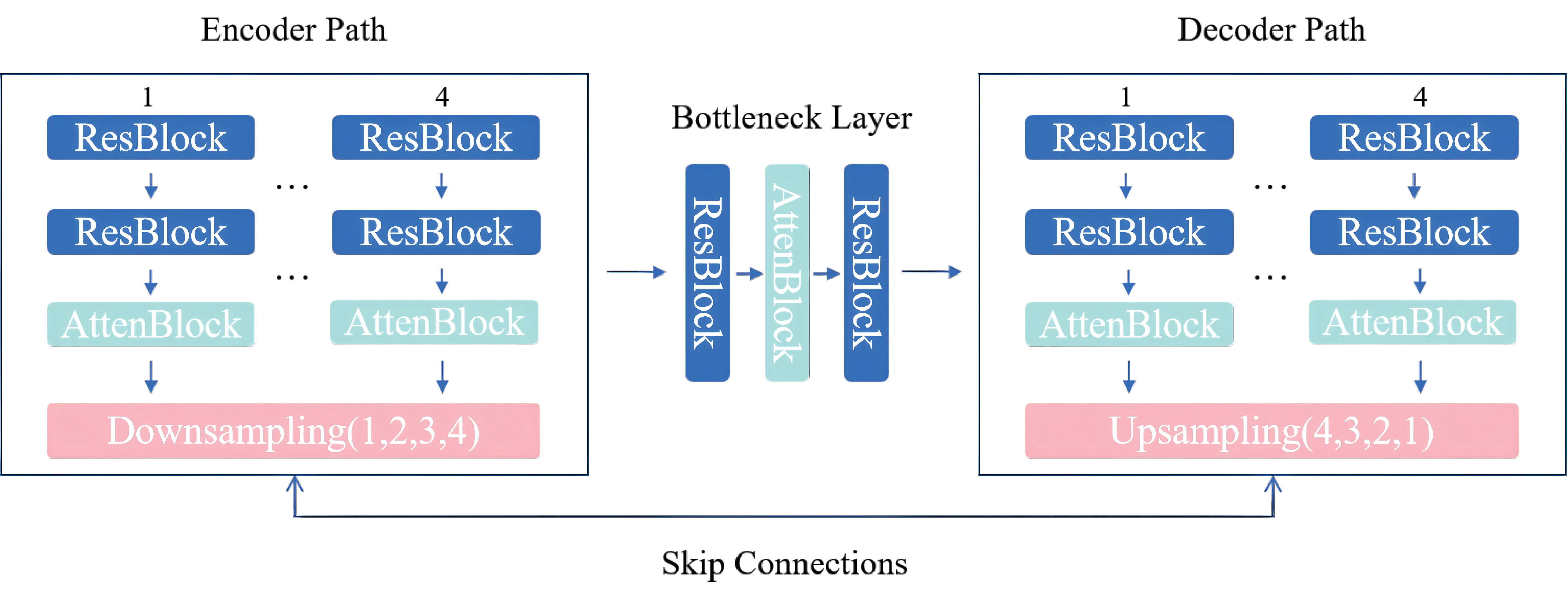}
    \caption{Convolution-based UNet architecture.}
    \label{fig:UNet}
    \vspace{-6mm}
\end{figure}
To adapt to neural network processing, the complex-valued CSI matrix $\mathbf{H}_t \in \mathbb{C}^{M \times N}$ is represented as the dual-channel real-valued tensor $\mathbf{X}_t \in \mathbb{R}^{2 \times M \times N}$, where the two channels correspond to the real and imaginary parts of the CSI, respectively. The input tensor $\mathbf{X}_t$ first passes through an initial convolution layer to adjust the number of channels before entering the encoder path. This path consists of 4 downsampling levels (1, 2, 3, 4). Each level contains 2 convolutional ResBlocks. AttenBlocks are inserted at specific resolutions to capture long-range dependencies. Simultaneously, upon completing the feature extraction at each level, the downsampling operation is performed to progressively reduce the spatial dimensions while expanding the number of channels. The outputs of each layer in the encoder are cached for subsequent skip connections.
\begin{algorithm}[h]
\setstretch{1.2}
    \caption{Training process for the proposed FM-based channel estimation}
    \label{alg:IVBL}
    \renewcommand{\algorithmicrequire}{\textbf{Input:}}
    \renewcommand{\algorithmicensure}{\textbf{Output:}}
    \begin{algorithmic}[1]
        \REQUIRE True MIMO channel dataset $\mathcal{H}=\{\mathbf{H}_{1}^{(i)}\}_{i=1}^{D}$, noise error matrix $\tilde{\mathbf{E}} \sim \mathcal{CN}(0, \tilde{\sigma}^2 \mathbf{I}_M)$, time step $t\in [0,1]$, 
        \STATE \textbf{Repeat}
       \STATE Generate noise channels matrix $\mathbf{H}_0$ via (7). 
       \STATE  Compute conditional flow mapping $\mathbf{H}_t$ via (12). 
        \STATE Offline train $u_t^\theta(\mathbf{H}_t)$ via (14).  
        \STATE \textbf{Until} \textbf{converged}
       \ENSURE Model parameters $u_{t}^{\theta}(\mathbf{H}_{t}).$
    \end{algorithmic}
\end{algorithm}

At the bottleneck layer, the model performs deep feature modelling at the lowest resolution through a combination of ResBlock and Attenblock to capture global contextual information. Subsequently, in the decoding path, the model progressively recovers the spatial resolution through upsampling operations. At each stage, features from the corresponding encoding layers are fused through skip connections. The design effectively balances semantic information extraction with spatial detail preservation.

The time step embedding vector $t$, serving as conditional information, is mapped to a high-dimensional feature via sinusoidal positional encoding and is then injected into every residual block, enabling fine-grained control over the generative process. Finally, the output layer maps the features back to the target channel dimension, producing the predicted velocity field $\mathbf{v}_t(\mathbf{H}_t | \mathbf{H}_1)$, which has the same dimensions as the input. By jointly optimizing across multiple resolution levels, the network can learn the complex spatial structural relationships and temporal evolution patterns inherent in the CSI matrix, thereby achieving accurate prediction of the channel velocity field. The training process of our proposed method is summarized in \textbf{Algorithm 1}.

\addtolength{\topmargin}{0.08in}
\begin{algorithm}[h]
\setstretch{1.2}
    \caption{Sampling process for the proposed FM-based channel estimation}
    \label{alg:IVBL}
    \renewcommand{\algorithmicrequire}{\textbf{Input:}}
    \renewcommand{\algorithmicensure}{\textbf{Output:}}
    \begin{algorithmic}[1]
        \REQUIRE Pilot matrix $\mathbf{P}$, measurement matrix $\mathbf{Y}$, pretrained flow matching model $u_{t}^{\theta}(\mathbf{H}_{t}) $, total sampling steps $S$. 
       \STATE  Initial channel estimation $\mathbf{H}_{est,0}$ via (16) 
       \STATE  Set the step size $\Delta t=1/S$.
       \FOR    {$s \leftarrow 1$ to $S$}   
       \STATE  Euler update ${{\mathbf{H}}_{est,s}}={{\mathbf{H}}_{est,s-1}}+s\cdot \Delta t\cdot u_{t}^{\theta }({{\mathbf{H}}_{est,s-1}})$ 
       \ENDFOR
       \ENSURE Estimated channel matrix $\mathbf{H}_{est}$
    \end{algorithmic}
\end{algorithm}
The sampling method based on SM-based model depends on annealed Langevin dynamics [16]. This process requires sequentially executing thousands of iterations, where each step moves a tiny step size $\varepsilon$ in the direction of the score function (i.e., the gradient of the log-probability density of the data) estimated by a neural network. This stochastic sampling mechanism results in slow sampling speed and high computational costs. In contrast, FM-based model directly learns a deterministic probability flow path. This path is driven by a velocity field \( u_{t}^{\theta}(\mathbf{H}_{t}) \) and is described by the ODE:
\begin{equation}
\frac{d}{dt} \mathbf{H}_{t} = u_{t}^{\theta}(\mathbf{H}_{t}).
\end{equation}
Because it adopts the linear interpolation path from optimal transport as its fundamental construction, i.e., \( \mathbf{H}_{t} = t \mathbf{H}_{1} + (1-t) \mathbf{H}_{0} \), this induces a velocity field \( v_{t}(\mathbf{H}_{t}) = \mathbf{H}_{1} - \mathbf{H}_{0} = -\hat{\mathbf{E}} \). This implies that the flow from \( \mathbf{H}_{0} \sim p_{0}(H_{0}) \) to \( \mathbf{H}_{1} \sim p_{1}(H_{1}) \) is ideally a uniform linear motion. In specific applications, the CSI matrix obtained via the easily implemented LS estimation is used as the initial state, i.e., $\mathbf{H}_{est,0}=\mathbf{H}_{LS}$,
\begin{equation}
\mathbf{H}_{LS} = \mathbf{P}^{H} \mathbf{Y} = \mathbf{H}_{1} + \hat{\mathbf{E}},
\end{equation}
where \( \hat{\mathbf{E}} = \mathbf{E} \mathbf{P}^{H} \sim \mathcal{CN}(0, \hat{\sigma }^{2} \mathbf{I}_{M}) \). The current velocity $ u_{t}^{\theta}(\mathbf{H}_{est,s-1})$ is calculated by the trained network. We set the step size \(\Delta t\) and update the channel with $S$ iterations using euler sampling,
\begin{equation}
{{\mathbf{H}}_{est,s}}={{\mathbf{H}}_{est,s-1}}+s\cdot \Delta t\cdot u_{t}^{\theta }({{\mathbf{H}}_{est,s-1}}).
\end{equation}
where $s\in [1,2,\cdots,S]$. Note that the solution trajectory of the ODE is approximately linear. This allows the step size \(\Delta t\) to be set relatively large, enabling the sampling phase to require only a few to a dozen iterations to achieve high estimation accuracy. Consequently, this approach significantly reduces computational overhead while maintaining generation quality. The sampling process of our proposed method is summarized in \textbf{Algorithm 2}.


\section{Numerical Results}
This section presents the numerical results for evaluating the proposed FM-based generative channel estimation scheme. We first introduce the experimental details, including the dataset, training and testing settings. Then, we conduct visualizations to demonstrate the effectiveness of the proposed scheme.

\subsection{Simulation Setup}
We employ the cluster delay line C (CDL-C) channel model defined by the 3rd Generation Partnership Project (3GPP) standard to generate true channel data [19]. This model is suitable for non-line-of-sight (NLOS) propagation conditions in urban macrocell environments. The carrier frequency is set to 40 GHz, the millimeter-wave (MmWave) band known for its large available bandwidth but also more significant path loss, which often necessitates the use of MIMO technology. The normalized distance between the antenna elements is set to 0.5. Two antenna configuration schemes: $(M, N) = (16, 64)$ and $(M, N) = (32, 128)$, are considered for the RF antennas at the transmitter and receiver.

We train the model on the dataset $\mathcal{H}$ comprising $D=10,000$ CDL-C channel samples, denoted as $\mathcal{H}=\{\mathbf{H}_{1}^{(i)}\}_{i=1}^{D}$, 1000 validation samples and evaluate its performance on a separate test set of 500 newly generated CDL-C channel realizations.
During the training phase, the model is trained for 100 epochs with the batch size of 128. We use the AdamW optimizer with the fixed learning rate of $10^{-5}$ and the weight decay of $10^{-2}$ to apply $L_2$ regularization, which helps improve the model's ability to learn useful features effectively.
During the testing phase, we simply perform euler sampling using the trained velocity field $u_{t}^{\theta }({{\mathbf{H}}_{t}})$, noting that the number of FM-based sampling steps in question is considerably lower compared to the SM-based method. Here we use the SM-based model, which completes its training in [14], where annealed Langevin dynamics performs $K=2311$ sampling steps and $L=3$ annealing updates are performed below each sampling step.
The estimation quality is quantified using the normalized mean square Error (NMSE), defined as:
\begin{equation}
\text{NMSE}[\text{dB}] = 10 \log_{10} \left( \frac{ \left\| \mathbf{H}_{\text{est}} - \mathbf{H}_{1} \right\|_{F}^{2} }{ \left\| \mathbf{H}_{1} \right\|_{F}^{2} } \right),
\label{eq:nmse_definition}
\end{equation}
where \(\mathbf{H}_{\text{est}}\) is the estimated channel matrix and \(\mathbf{H}_{1}\) is the true channel matrix.

\begin{figure}
    \centering
    \includegraphics[width=0.76\linewidth]{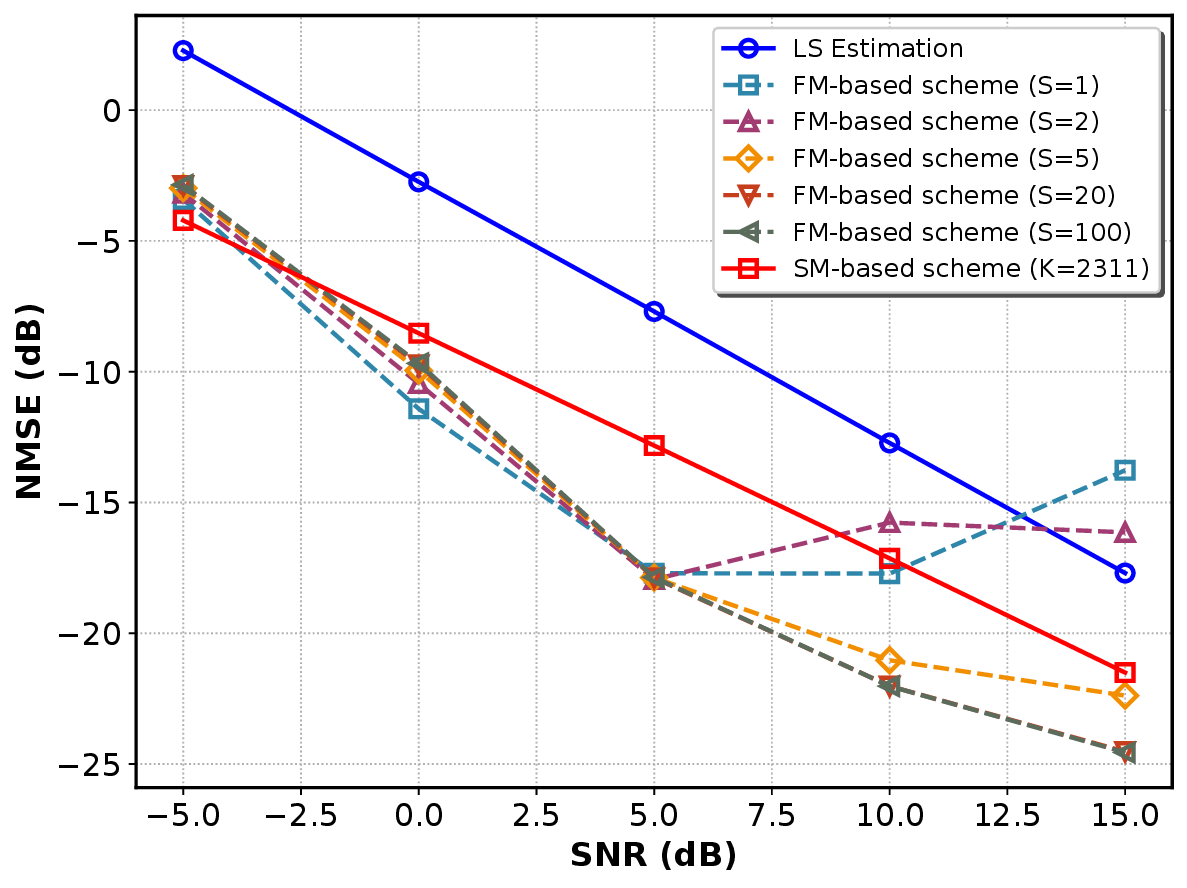}
    \caption{Comparison of NMSE performance for various SNRs with length of pilot $T=75$, M=16, N=64, and the noise parameter $\tilde{\sigma}=0.1$.}
    \label{fig:enter-label}
    \vspace{-3mm}
\end{figure}
\subsection{Performance Evaluation of the Proposed FM-based Channel Estimation}
As shown in Fig. 2, the NMSE of the FM-based scheme, the LS scheme, and the SM-based scheme decrease as the SNR increases. Due to its sensitivity to noise, the LS estimation performs significantly worse in the low-SNR region. The proposed FM-based method uses LS estimation as the initial value and performs iterative optimization to obtain better estimation performance. When only $S=1$ is used, the discretization is too coarse to follow the optimal probability flow, leading to noticeable degradation at medium and high SNR. Increasing $S$ to 2, 5, and 20 progressively improves the NMSE, confirming that finer discretization better approximates the underlying continuous flow dynamics. Compared to the SM-based approach, FM achieves comparable or even better NMSE at a significantly lower number of iterations.
Although the SM-based scheme performs moderately well, but requires $K=2311$ inverse iterations, which is an extremely high computational overhead

\begin{figure}
    \centering
    \includegraphics[width=0.76\linewidth]{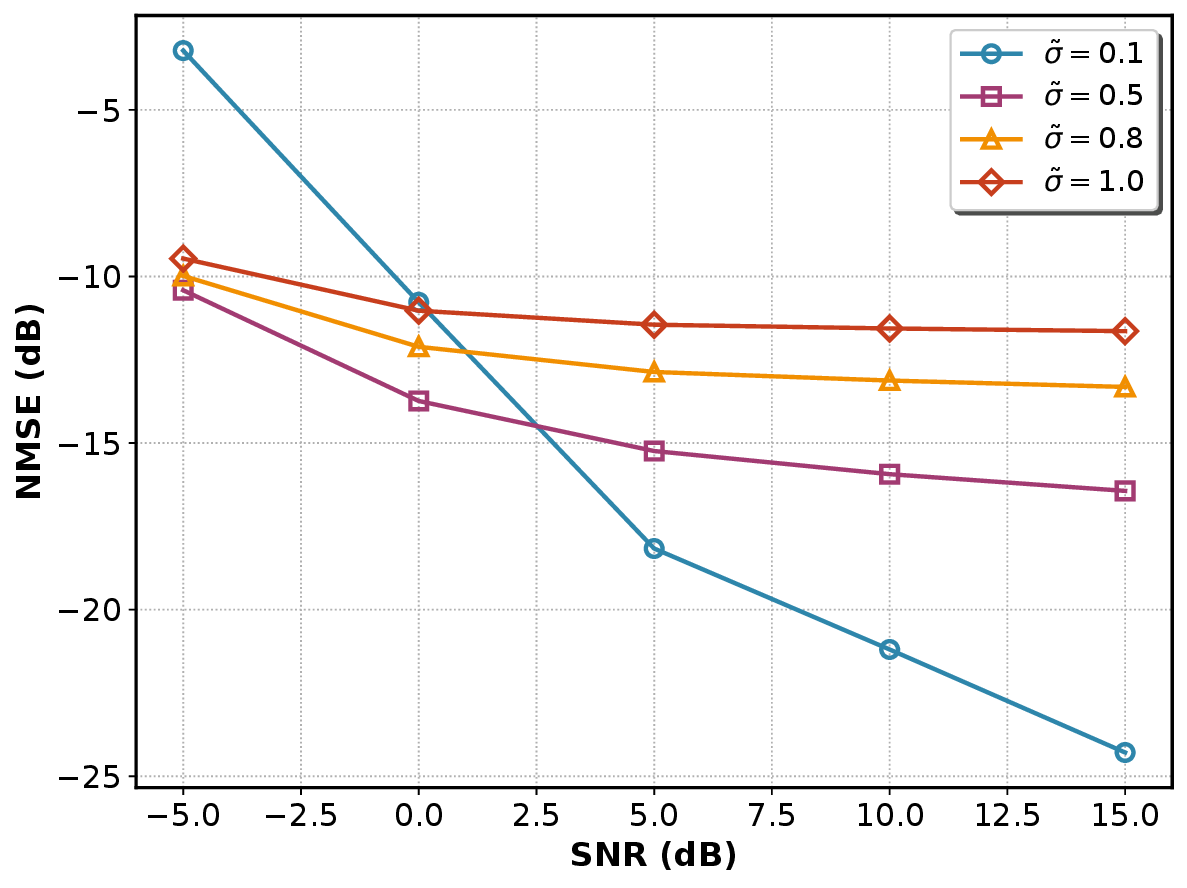}   
    \caption{Comparison of NMSE of FM-based method at various the noise parameter $\tilde{\sigma}$ with pilot length $T = 75$, and M=16, N=64.}
    \label{fig:enter-label}
    \vspace{-5mm}
\end{figure}
Fig. 3 illustrates the performance curves of NMSE with SNR under different training noise levels $\tilde{\sigma }$. The most striking overall phenomenon is the monotonically decreasing trend of NMSE under all curves as SNR increases. We observe the steepness of the curve's descent to notice that the curve corresponding to the small $\tilde{\sigma }=0.1$ has its NMSE decreasing faster with increasing SNR. This is because the noisy channel distribution and the true channel distribution are geometrically very close to each other, and the training model obtains an accurate uniform straight-line path that effectively steers the estimates to the true channel even from a highly noisy starting point at low SNR. The larger $\tilde{\sigma }$ , which learns that the velocity field deviates from the straight line assumption, this is more obvious at high SNR. However, the higher robustness at low SNR is due to the fact that the model has already seen very strong noise during training. Thus, the $\tilde{\sigma }$ should weigh the settings in a practical situation.
\begin{table}[h]
\centering
\small 
\caption{Sampling time comparison of different methods }
\label{tab:sampling_time}
\begin{tabular}{lccccc} 
\toprule
\textbf{Method} & \textbf{SM} & \multicolumn{4}{c}{\textbf{FM (Number of Steps)}} \\ 
\cmidrule(lr){3-6} 
Sampling Steps &2311 & 1 & 5 & 20 & 100 \\ 
\midrule
Sampling Time (s) & 91.444 & 0.069 & 0.340 & 1.363 & 6.825 \\
\bottomrule
\end{tabular}
\end{table}

As shown in TABLE \ref{tab:sampling_time}, we perform channel estimation for 500 data points at SNR=10dB, $(M, N) = (16, 64)$. The results indicate that the SM-based scheme requires the considerable number of sampling steps (2311 steps) to achieve notable performance, with a total time consumption of up to 91.444s. It is worth noting that the SM-based scheme requires only fewer sampling steps to achieve better performance than the SM-based scheme. Even if the number of FM-based method sampling steps increases to 100 (taking 6.825s), it is still more than 13 times faster than the SM-based method. This result strongly demonstrates that the FM-based methed can greatly reduce computational overhead while guaranteeing performance and meeting the demanding requirements for low latency in future communication systems.

\begin{figure}
    \centering
    \includegraphics[width=0.76\linewidth]{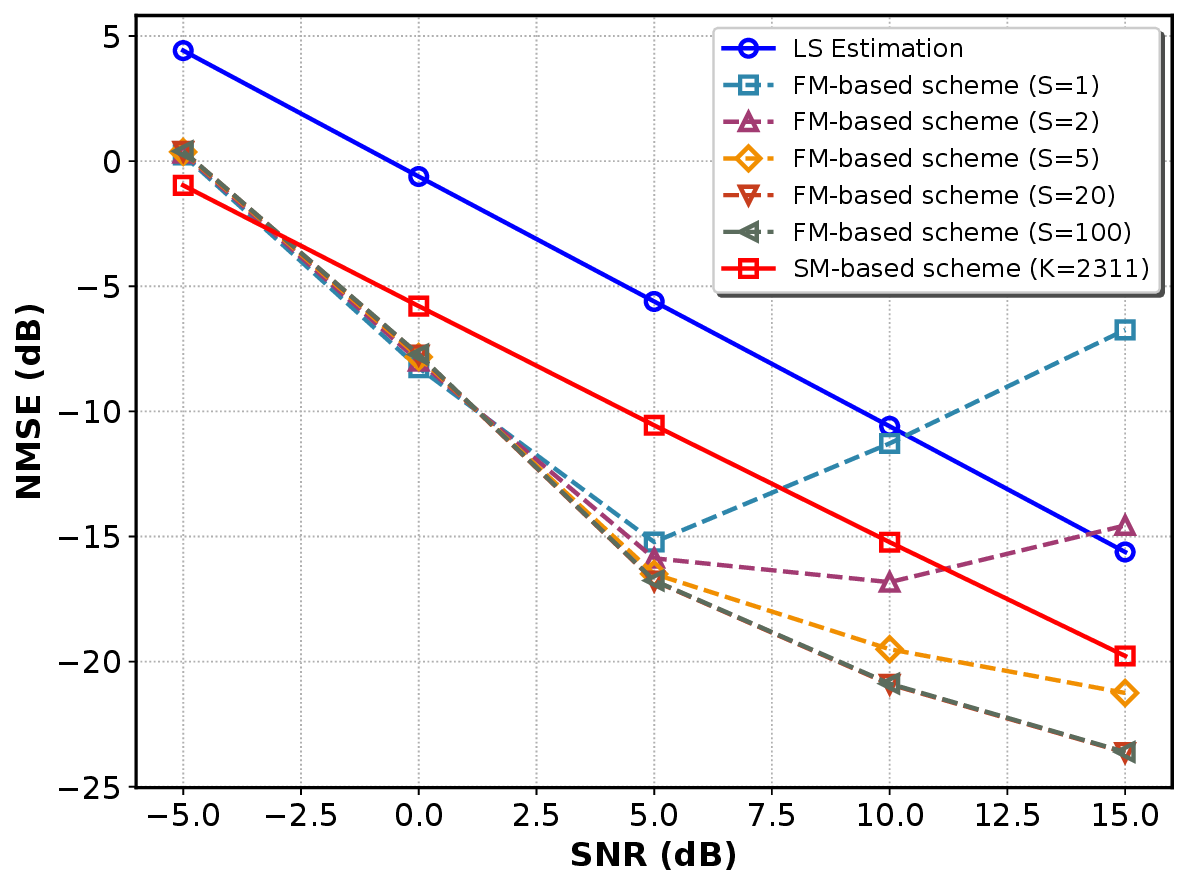}
    \caption{Comparison of NMSE performance for various SNRs with length of pilot $T=145$, M=32, N=128, and the noise parameter $\tilde{\sigma}=0.1$.}
    \label{fig:enter-label}
    \vspace{-4mm}
\end{figure}
As shown in Fig. 4 presents the performance comparison over different SNR ranges in a the MIMO scenario when the pilot length T=145 and the antenna configurations are M=32 and N=128. 
The experimental results show that the proposed FM-based scheme outperforms the conventional LS estimation and SM-based scheme despite the increase in channel dimension. Note that when the sampling step size is greater than or equal to 5, it achieves well estimation performance, which validates the advantage of the proposed scheme in balancing accuracy and efficiency in high-dimensional channel estimation tasks.
\section{CONCLUSION}
This paper introduces the FM-based scheme for MIMO channel estimation that leverages the denoising ability of FM to achieve accurate CSI. 
By formulating the channel estimation problem within FM framework, the conditional probability path is constructed from the noisy channel distribution to the true channel distribution. 
The core advantage of the proposed method lies in its deterministic probabilistic flow ODE path, which is designed based on the principle of linear interpolation in the optimal transmission theory, enabling the sampling process to converge quickly with a straight-line trajectory.
Experimental results demonstrate that our proposed scheme only needs 1-20 iterations to saturate the performance and achieve high-quality channel estimation with significantly reduced computational overhead.

\section{Acknowledgment}
This work was supported by the joint project of China Mobile Research Institute \& X-NET.

\vspace{12pt}
\end{document}